\documentclass[conference]{IEEEtran}
\IEEEoverridecommandlockouts
\usepackage{cite}
\usepackage{amsmath,amssymb,amsfonts}
\usepackage{algorithmic}
\usepackage{graphicx}
\usepackage{textcomp}
\usepackage{xcolor}
\usepackage{tabularx}
\usepackage{url}
\usepackage{tikz}
\usepackage{amsmath}
\usetikzlibrary{positioning,calc}

\def\BibTeX{{\rm B\kern-.05em{\sc i\kern-.025em b}\kern-.08em
    T\kern-.1667em\lower.7ex\hbox{E}\kern-.125emX}}
\begin{document}

\title{PRISM-Consult: A Panel-of-Experts Architecture for Clinician-Aligned Diagnosis\\
}

\author{
\IEEEauthorblockN{Lionel Levine, PhD}
\IEEEauthorblockA{\textit{University of California, Los Angeles}\\
Los Angeles, CA 90095, USA\\%
ORCID: 0000-0002-6926-7438}
\and
\IEEEauthorblockN{John Santerre, PhD}
\IEEEauthorblockA{\textit{UC Berkeley School of Information}\\
Berkeley, CA 94720, USA\\
(ORCID not available)}
\and
\IEEEauthorblockN{Alexander S. Young, MD MSHS}
\IEEEauthorblockA{\textit{UCLA David Geffen School of Medicine}\\
Los Angeles, CA 90095, USA\\
ORCID: 0000-0002-9367-9213}
\and
\IEEEauthorblockN{T. Barry Levine, MD}
\IEEEauthorblockA{\textit{ABLE Medical Consulting}\\
Savannah, GA 31411, USA\\
(ORCID not available)}
\and
\IEEEauthorblockN{Francis Campion, MD}
\IEEEauthorblockA{\textit{MITRE Corporation}\\
Bedford, MA 01730, USA\\
ORCID: 0000-0002-0757-9305}
\and
\IEEEauthorblockN{Majid Sarrafzadeh, PhD}
\IEEEauthorblockA{\textit{University of California, Los Angeles}\\
Los Angeles, CA 90095, USA\\
(ORCID not available)}
}

\maketitle

\begin{abstract}
We present PRISM-Consult, a clinician-aligned panel-of-experts architecture that extends the compact PRISM sequence model into a routed family of domain specialists. Episodes are tokenized as structured clinical events; a light-weight router reads the first few tokens and dispatches to specialist models (Cardiac–Vascular, Pulmonary, Gastro–Oesophageal, Musculoskeletal, Psychogenic). Each specialist inherits PRISM’s small transformer backbone and token template, enabling parameter efficiency and interpretability. This initial study evaluates a scoped panel of five specialist families defined by high-impact ED diagnostic groups. On real-world Emergency Department cohorts, specialists exhibit smooth convergence with low development perplexities across domains, while the router achieves high routing quality and large compute savings versus consult-all under a safety-first policy. We detail the data methodology (initial vs.\ conclusive ICD-9 families), routing thresholds and calibration, and report per-domain results to avoid dominance by common events. The framework provides a practical path to safe, auditable, and low-latency consult at scale, and we outline validation steps—external/temporal replication, asymmetric life-threat thresholds, and multi-label arbitration—to meet prospective clinical deployment standards.

\end{abstract}

\begin{IEEEkeywords}
Clinical decision support, Emergency department, Triage, Healthcare AI, Large language models, Transformers, Mixture-of-experts, Routing, Probability calibration, Natural language processing, Tokenization, ICD-9-CM, Multi-label classification, Real-time systems, Safety-critical systems, Model interpretability, Electronic health records
\end{IEEEkeywords}


\section{Background}
Clinical care unfolds as time-ordered sequences of events that include presenting complaints, assessing vitals, diagnostic orders, laboratory observations, and evolving diagnoses. The core challenge is to turn the earliest, sparsest portion of this sequence into reliable predictions that (1) anticipate likely next events and differentials and (2) support time-critical decision-making. Our goal is to achieve this with models that are fast, auditable, and robust to noisy early signals.
In the Emergency Department (ED), where minutes matter and presentations are often ambiguous, we represent encounters as tokenized event streams and learn to predict the next event and, by extension, the patient’s diagnostic trajectory. 

We previously introduced (\emph{PRISM}) a compact, domain-focused approach to sequence modeling\cite{levine2025prism}. We focused on the ED setting as a proving ground for sequence-based clinical decision support, emphasizing early-prefix prediction that can surface high-value differentials and guide downstream diagnostic actions. PRISM represented Emergency Department (ED) episodes as tokenized event streams and trained a small decoder-only transformer to predict the next event and, by extension, the evolving diagnostic trajectory \cite{levine2025prism}. PRISM emphasized \emph{low-latency, interpretable modeling} by (i) restricting the event schema to diagnostics, labs, and diagnoses (excluding procedures and medications), (ii) employing a hand-designed token template (e.g., \texttt{[DIAG]\_ICD9\_410.xx}, \texttt{[OBS]\_LAB\_TROP:HIGH}, \texttt{[ACTION]\_ORD\_ECG}), (iii) capping sequence length (e.g., 512 tokens) and vocabulary size (compact custom lexicon), and (iv) utilizing a moderate-capacity backbone (e.g., 6 layers, $d_{model}{=}256$) suitable for clinical deployment constraints.

Initially applied to chestpain presentations in the ED, PRISM demonstrated strong calibration and efficient inference, validating the premise that carefully scoped tokenization and model size can yield clinically useful predictions without the cost or opacity of very large general-purpose models. However, the same design choices that improve efficiency, namely a streamlined vocabulary and a single-target clinical domain, also \emph{narrow} the applicability of the model: real-world emergency department presentations often traverse multiple organ systems, and differential diagnoses for a single symptom (e.g., chest pain) span cardiac, pulmonary, gastro-oesophageal, musculoskeletal, and psychogenic causes. This motivates an extension from a single-domain backbone to a \emph{routed panel of domain specialists}.

\section{INTRODUCTION}

\subsection{Clinical motivation and use case}
In the Emergency Department (ED), clinicians must make time-critical decisions from sparse, noisy early signals: a chief complaint, a small number of initial vitals, and the first diagnostic actions (e.g., ECG, labs, imaging orders). At this stage, the task is rarely ``predict the final diagnosis'' in a single step; instead, clinicians form an evolving differential and decide which actions are most urgent and informative. This creates a practical need for \emph{early} decision support that can suggest likely next steps and plausible differentials while remaining fast, auditable, and conservative under uncertainty.

\subsection{Why clinician-aligned routing}
Real ED care is organized around consult services and organ-system expertise (e.g., cardiology, pulmonary/critical care, gastroenterology). We therefore frame early prediction as a \emph{routing} problem: from the earliest events, decide which specialist perspectives are most appropriate to engage. This design supports safety-first behavior (e.g., consult multiple specialists when uncertain) and enables explicit control over compute/latency trade-offs while preserving clinical interpretability: a routed specialist corresponds to a clinically recognizable ``who should weigh in?'' decision.

\subsection{Why transformer specialists rather than a single monolith}
A single sequence model can be effective, but ED trajectories often cross domains and contain heterogeneous patterns (e.g., dyspnea spanning pulmonary embolism, pneumonia, heart failure, anxiety). Specialization allows each model to learn domain-consistent thresholds, priors, and event semantics while retaining a shared tokenization template and backbone for parameter efficiency. The resulting panel improves coverage without requiring an ever-expanding single-model vocabulary, and it allows domain-specific calibration and evaluation aligned to clinical consequences.

In this paper we present \textbf{PRISM-Consult}, a routed \emph{panel-of-experts} architecture that extends the PRISM methodology from a single-domain model to a family of specialist models orchestrated by a lightweight router. The router interprets the earliest events (symptoms and first diagnostic cues) and dispatches the episode to one or more specialist models aligned with clinical organ systems. Each specialist inherits the PRISM tokenization template, shares embeddings for consistency, and is fine-tuned on domain-filtered cohorts labeled by definitive ICD-9 families. This preserves PRISM’s efficiency and interpretability while expanding clinical coverage.

Our contributions are threefold. First, we formalize a \emph{clinical routing surface} based on initial ICD-9 symptom codes and map it to specialist targets defined by conclusive diagnostic families. Second, we implement a minimalist router and parameter-efficient specialist adapters that retain PRISM’s compact footprint while providing cross-domain accuracy. Third, we report partial results across cardiac, pulmonary, gastro-oesophageal, musculoskeletal, and psychogenic domains, showing smooth convergence and low development perplexities (near $\sim$2.0 for all five completed specialists), thereby validating that the PRISM backbone generalizes when routed to domain-consistent corpora.

\subsection{Why a Panel of Experts? Tokenization Trade-offs and Clinical Coverage}
Its tokenization strategy sits at the core of PRISM’s efficiency. A \emph{more specific} token inventory (e.g., separate tokens for finely binned labs, highly granular diagnoses) captures nuanced patterns and can improve in-domain accuracy and calibration. Yet specificity increases vocabulary size, fragmenting data across rare tokens and slowing inference. Conversely, a \emph{more generic} inventory (broader bins, fewer code variants) compacts the vocabulary and speeds inference but risks losing discriminative power across heterogeneous diseases.

This creates a fundamental tension in single-model designs: optimizing the token schema for one domain (e.g., cardiology) may degrade performance in another (e.g., pulmonary embolism vs.\ pneumothorax), where different signals and thresholds matter. 

\subsection{Heterogeneity, Comorbidity, and Cross-Specialty Trajectories in the ED}

Further complicating diagnostic predictions is the fact that emergency presentations are intrinsically heterogeneous and frequently comorbid: a single complaint (e.g., chest pain, dyspnea, syncope) often admits plausible explanations across multiple organ systems, and early measurements (vitals, first labs, first orders) carry overlapping, sometimes contradictory signals. In MIMIC-IV ED cohorts, the same prefix of events can legitimately evolve toward cardiac (AMI, dissection), pulmonary (PE, pneumothorax, pneumonia), gastro–oesophageal (Boerhaave, spasm, reflux), musculoskeletal, or psychogenic end points, or some combination thereof, and patients routinely carry chronic conditions (CKD, COPD, diabetes, atrial fibrillation) that shift priors and confound intermediate tests. These realities create diagnostic pathways that traverse traditional service boundaries and unfold via cross-specialty handoffs. 

This cross-disciplinary drift introduces several modeling hazards for single-domain specialists. First, \emph{boundary undercoverage}: a specialist tuned to one domain’s vocabulary and thresholds may systematically down-weight rare but life-threatening out-of-domain hypotheses that share early features (e.g., PE vs.\ AMI in troponin-positive dyspnea). Second, \emph{calibration shift}: domain-specific likelihoods calibrated in-domain can become miscalibrated when the latent etiology changes mid-episode, especially under sparse, prefix-only supervision. Third, \emph{premature closure}: strong local patterns (e.g., ECG orders $\rightarrow$ ischemia) can anchor the trajectory before counter-signals appear (e.g., pleuritic descriptors, D-dimer, CTA), yielding myopic next-event proposals.

Sequential structure further compounds these risks. Clinical events arrive as partially ordered sets (co-occurring orders, batched lab panels) with irregular time gaps, and many ED timelines are better viewed as \emph{sequences of sets} than pure token strings. Methods that enforce set-level invariance or augment code streams with ontology signal (e.g., CCS/knowledge-graph embeddings) help preserve meaning across co-occurrences and synonyms, but they do not by themselves resolve cross-service drift when the \emph{generator} of observations changes during workup\cite{dpss_2020} \cite{setor_2023}. Likewise, zero-shot generative rollouts over patient-health timelines demonstrate breadth but still reflect the chosen pretraining mix and may blur service-specific decision thresholds \cite{ethos_2024}.

From a probabilistic standpoint, the ED setting resembles large-context inference over many interacting latent causes. Classic graphical approaches (e.g., QMR-DT) formalize comorbidity and conditional dependence but become computationally brittle at realistic scales \cite{jaakkola1999variational}. Autoregressive transformers approximate these high-order conditionals efficiently, yet as already noted above, a \emph{single} backbone must compromise between compact vocabularies (speed, stability) and domain-specific expressivity (rare tokens, narrow thresholds). In practice, optimizing tokenization and priors for one service can degrade discrimination in another, particularly near safety-critical boundaries.

These observations motivate a \emph{routed, multi-disciplinary engagement} approach. Concretely it: (i) reads the earliest prefix and estimate calibrated probabilities over clinical domains; (ii) consults one or more domain specialists when warranted (with asymmetric, safety-first thresholds for life-threatening etiologies); (iii) arbitrates multi-label suggestions with deterministic, auditable policies; and (iv) fails open when confidence is low or vitals breach guardrails. Such conditional computation aligns with mixture-of-experts and expert-routing advances—activating only the parameters relevant to the emerging hypothesis space while preserving latency and traceability.

\textbf{PRISM-Consult} is designed with this approach in mind. By \emph{holding the token template constant}—preserving interpretability and shared embeddings—while \emph{routing} episodes to \emph{domain-specialized fine-tunings}. Each specialist selectively expands or emphasizes a \emph{local} subset of tokens relevant to its domain (e.g., pleural processes, gas exchange markers for pulmonary; ischemic markers and ECG patterns for cardiac), without forcing the global vocabulary to balloon. The router thus acts as a clinical analogue of referral: ``right expert, right time,'' mitigating the specificity–coverage trade-off inherent to a single global model.

\section{Related Work}
\subsection{Diagnostic Progressions and LLM Token Prediction}

In large language models (LLMs), each token—representing a word or sub-word unit—has clear pairwise probabilities relative to other tokens. For example, given a token such as "New," the subsequent token might have high probability for "York," "Zealand," or "Jersey." Individually, these pairwise probabilities are straightforward. However, as the sequence of tokens expands—considering an entire preceding sentence or paragraph—the combinatorial complexity and contextual dependencies make it impossible to precisely calculate probabilities using first order correlations alone. Hence, deep learning and attention mechanisms in transformer architectures (like GPT) approximate these complex conditional dependencies, capturing long-range contexts and intricate relationships between tokens.

By analogy, diagnostic medicine faces a similar complexity. Individual diagnostic tests may exhibit well-defined pairwise probabilities for specific diseases—just as pairwise token probabilities can be well-established. However, when attempting to predict the next best test or disease diagnosis given the entire preceding sequence of events (the patient's complete history, including medical tests, symptoms, prior diagnoses, lifestyle factors, and comorbidities), the scenario becomes profoundly more complex. The interactions between these elements are intricate, context-dependent, and frequently nonlinear, making exact probabilistic inference practically infeasible.

Thus, Bayesian statistical approaches quickly reach limitations. Transformer-based diagnostic models, analogous to language models, can approximate the complex contextual dependencies inherent in patient histories, making them potentially more capable of accurately predicting subsequent medical events (tests, results, or diagnoses). These models implicitly learn from data to reflect patient-level complexity more naturally, offering both predictive capability and interpretable attention scores to highlight influential factors in diagnostic reasoning.

\subsection{Prior Work in Transformer and Deep Learning Models for Sequential Clinical Diagnosis}


In contrast to high-level summations of pre-existing efforts, which are considerable and highly varied in nature, we opt to focus in on prior work of immediate relevance to our proposed approach, by zeroing in on a methodology-specific approaches that have informed our own experimental design, and serve as a basis for which our work attempts to build on. 

\subsubsection{Tokenization and Representation of Medical Data}
\label{sec:tokenization}

The first step in adapting transformer models to diagnostic-sequence prediction is
to convert a patient’s richly structured history into a sequence of discrete
\emph{tokens}.  Early work relied on standardised vocabularies such as
ICD-9/ICD-10, CPT and Read codes, treating each code as an individual token
\cite{medalbert_2024,lotusai_2025}.  A coarser alternative aggregates thousands
of raw codes into a few hundred clinically coherent groups, reducing sequence
length while retaining clinical meaning \cite{medalbert_2024}.

More granular pipelines tokenise free-text concepts, symptoms and history
fragments extracted from clinical narratives \cite{autoddx_2024}.  The ETHOS framework extend this idea further by encompassing: admissions, diagnoses (ICD-10-CM), procedures (ICD-10-PCS),
medications (ATC), laboratory results (LOINC) and even inter-event time gaps. Under the ETHOS framework, these events are
mapped to 1–7 tokens each, then ordered chronologically into a patient-health
timeline (PHT) \cite{ethos_2024}.

Because multiple events can occur simultaneously during a visit, several authors
represent a patient record as a \emph{sequence of sets}, forcing the model to be
order-invariant within each set.  DPSS is the canonical example of this strategy
\cite{dpss_2020}.  Finally, semantic enrichment with external knowledge graphs
(or hierarchical ontologies such as CCS) can be injected by concatenating
concept embeddings with ontology embeddings, as done in the SETOR framework
\cite{setor_2023}.

\subsubsection{Transformer Model Architectures in Diagnostic Prediction}
\label{sec:architectures}

Traditionally model choice varied depending on the prediction task.  Encoder-only families (BERT,
ALBERT, MedAlbert) excel at classification problems such as early disease
detection from longitudinal code streams; MedAlbert, for instance, uses a
6-layer ALBERT encoder to flag incipient lung cancer three years in advance
\cite{medalbert_2024}.  Decoder-only designs (GPT-like) suit generative
forecasting: ETHOS, for instance, employs a causal decoder to roll out future PHTs in a zero-shot
setting \cite{ethos_2024}.  Hybrid encoder–decoder stacks (e.g.,
TransformEHR) translate past encounters into predicted future ICD sequences
\cite{lotusai_2025}.

Domain-aware modifications are common.  SETOR feeds visit-level embeddings—
augmented with CCS-derived ontology vectors and continuous-time positional
encodings—into a multi-layer transformer encoder dubbed the \emph{Patient
Journey Transformer} \cite{setor_2023}.  Such customisations can improve both
performance and interpretability without abandoning the core attention
machinery of the original architecture.

\subsection{Additional Clinical foundation models and domain adaptation}
Foundation models trained on biomedical and clinical corpora underpin many recent advances in clinical NLP. Early domain-adapted encoders such as BioBERT~\cite{lee2019biobert} and ClinicalBERT~\cite{huang2019clinicalbert} demonstrated that continued pretraining on PubMed/PMC and EHR notes yields substantial gains for NER, RE, QA, and risk prediction. PubMedBERT~\cite{gu2021pubmedbert} further showed that training \emph{from scratch} on biomedical text can outperform continual-pretraining approaches on multiple benchmarks. Generative clinical LMs have since emerged: BioGPT~\cite{luo2022biogpt} brought GPT-style pretraining to biomedical literature, while GatorTron and GatorTronGPT scaled to billions of parameters using mixed corpora that include large EHR collections, improving a wide array of downstream tasks~\cite{yang2022gatortron,peng2023gatortrongpt}. Complementing these efforts, the Med-PaLM line explored instruction tuning and medical QA evaluation at scale, reporting expert-competitive answers on MultiMedQA and related evaluations~\cite{singhal2023largelanguage,medpalm2}.

Despite strong aggregate performance, deploying a single generalist model for all problems can be suboptimal in clinical operations. Differences in vocabulary, labeling practices, and decision latencies across services (e.g., cardiology vs.\ psychiatry) motivate domain-aligned specializations with small, auditable backbones, provided we can direct cases reliably to the right specialist. This motivates the \emph{routed panel-of-experts} framing adopted in PRISM-Consult.

\subsection{Mixture-of-experts and routing among specialists}
Mixture-of-Experts (MoE) and conditional computation architectures enable input-dependent parameter activation and have been central to efficient scaling. GShard~\cite{lepikhin2020gshard} and Switch Transformer~\cite{fedus2021switch} demonstrated that sparse expert routing can train trillion-parameter class models with near-constant per-token cost, provided stability and load-balancing constraints are met. Subsequent work proposed alternative routing rules (e.g., Expert Choice) to improve load balancing and efficiency~\cite{expertchoice}. In parallel, “routing among models” at inference---choosing which pre-trained system to run for a given query---has become an active area, with routers trained from preference data or heuristics to trade off cost, quality, and latency~\cite{routellm-openreview,eagle-router}.

Beyond general LLM systems, ensemble/agent-style methods explicitly coordinate multiple specialists. Recent “mixture-of-agents” approaches show that structured collaboration among models can improve reasoning and robustness~\cite{moa2024}. In healthcare, most prior work focuses on single-model evaluation (e.g., LLMs for triage or diagnosis)~\cite{masanneck2024triage,jama2024diagnostic} or traditional clinical decision support (CDS)~\cite{wasylewicz2018cds}. There remains a gap for lightweight, auditable routers that triage \emph{early} clinical prefixes to compact, domain-specific experts, optimizing safety and compute while preserving traceability—precisely the niche that PRISM-Consult targets.

\subsection{Clinical sequence similarity and pathway clustering}
A long line of work in clinical sequential pattern mining and pathway analysis focuses on defining similarity between patient trajectories and clustering them into clinically meaningful pathways. Temporal similarity measures are central in this setting because clinical workflows depend not only on event identity but also on ordering, repetition, and time gaps \cite{combi2009temporal}. Pathway clustering methods adapt classical alignment/edit-distance algorithms (e.g., Needleman--Wunsch variants) to group patient journeys into interpretable clusters \cite{zhang2015cowpath}. Related studies also examine clustering diagnosis-code sequences jointly with demographics to discover stratified pathway structure \cite{zhong2022clustering}. These approaches primarily target cohort discovery, summarization, and retrospective pathway characterization; by contrast, PRISM-Consult targets \emph{prospective early-prefix prediction} and clinician-aligned dispatch to specialist models. Nonetheless, similarity-based pathway work informs what constitutes a ``near miss'' versus a ``severe miss,'' motivating our cost-sensitive evaluation.

\subsection{Probabilistic sequence models and stochastic automata}
Classical probabilistic approaches such as Markov models, hidden Markov models, and stochastic automata have been applied to encoded clinical sequences due to their interpretability and tractability \cite{budde2017road}. These methods can perform well when the state space is limited and the temporal process is relatively stationary. However, ED event streams are heterogeneous and high-dimensional, with long-range dependencies and sparse early prefixes that are difficult to capture with low-order Markov assumptions without aggressive state aggregation. We therefore focus on compact transformer specialists with shared tokenization, while explicitly comparing against and discussing probabilistic alternatives where appropriate.

\subsection{Positioning PRISM-Consult}
PRISM-Consult operationalizes a routed specialist architecture tailored to ED presentations: (i) compact specialists (same PRISM backbone and schema) trained on domain-consistent corpora; (ii) a calibrated router that reads the first $K$ tokens and dispatches to one or more experts under safety-first thresholds; and (iii) per-domain evaluation and macro-averaging to avoid dominance by common presentations. Compared with monolithic clinical LMs, this design aligns with operational realities (service-specific vocabulary and SLAs), supports explainability (specialist audit trails and calibrated probabilities), and can scale horizontally by adding experts or refining routing policies, drawing on MoE and LLM-routing principles while remaining deployable in resource-limited clinical settings.

\section{Methods}

\subsection{Overview}
PRISM-Consult extends the original PRISM framework---a compact, domain-focused sequence model over structured clinical events---into a routed, multi-specialist system. All data handling, event schema design, tokenization, and baseline modeling choices follow PRISM; below we restate those elements in full and then describe the extensions for the router and specialist models. Throughout, we reference the prior work for provenance \cite{levine2025prism} while ensuring this section is self-contained.


\subsection{Data and Preprocessing}
\subsubsection{Data Source}
Data for this study was extracted from the MIMIC-IV database, an extensive electronic health records repository collected from patients admitted to the Beth Israel Deaconess Medical Center (BIDMC) between 2008 and 2019. The database encompasses detailed clinical information, including demographics, vital signs, laboratory test results, diagnostic procedures, and discharge diagnoses captured during hospital stays \cite{johnson2023mimiciv}.

\subsubsection{Data structure and cohort window}
Each record for this model is comprised of a comprehensive longitudinal record of a patient's diagnostic journey, across clinical episodes. Episodes begin at admission and end at patient discharge.

Timelines consist of timestamped \emph{structured} events drawn from:
\begin{enumerate}
    \item \textbf{Patient Information} (E.g., age, gender)
    \item \textbf{Admission and Discharge Events}
  \item \textbf{Diagnostics and orders} (e.g., ECG ordered, CTA chest ordered).
  \item \textbf{Laboratory observations} (e.g., troponin value, D-dimer positive/negative).
  \item \textbf{Diagnoses} encoded in ICD-9-CM.
\end{enumerate}
Procedures and medication administrations are \emph{intentionally excluded}, mirroring PRISM's emphasis on a concise and clinically interpretable vocabulary focused on diagnostic reasoning. All events are normalized to a patient-relative clock (Initial admission~$=0$) and sorted chronologically and then in a structured manner detailed below to ensure consistency in batched event tokenization.

\subsubsection{Inclusion/exclusion Criteria}
Our study included all adult ED encounters with at least one qualifying initial (pre-diagnosis) symptom code recorded during the index window, along with at least one of the identified terminal diagnoses, in order to have a comprehensive set of patients traversing the diagnostic progression across clinical disciplines.

\subsubsection{Initial (Pre-Diagnosis) ICD-9-CM Code Set — Router Inputs}
Figure 1 details the set of preliminary diagnostic codes a patient presents with that are notionally indicative of a diagnosis pathway. Inclusion of at least one such code is a requisite for inclusion in the study.
\begin{figure}
    \centering
    \includegraphics[width=1\linewidth]{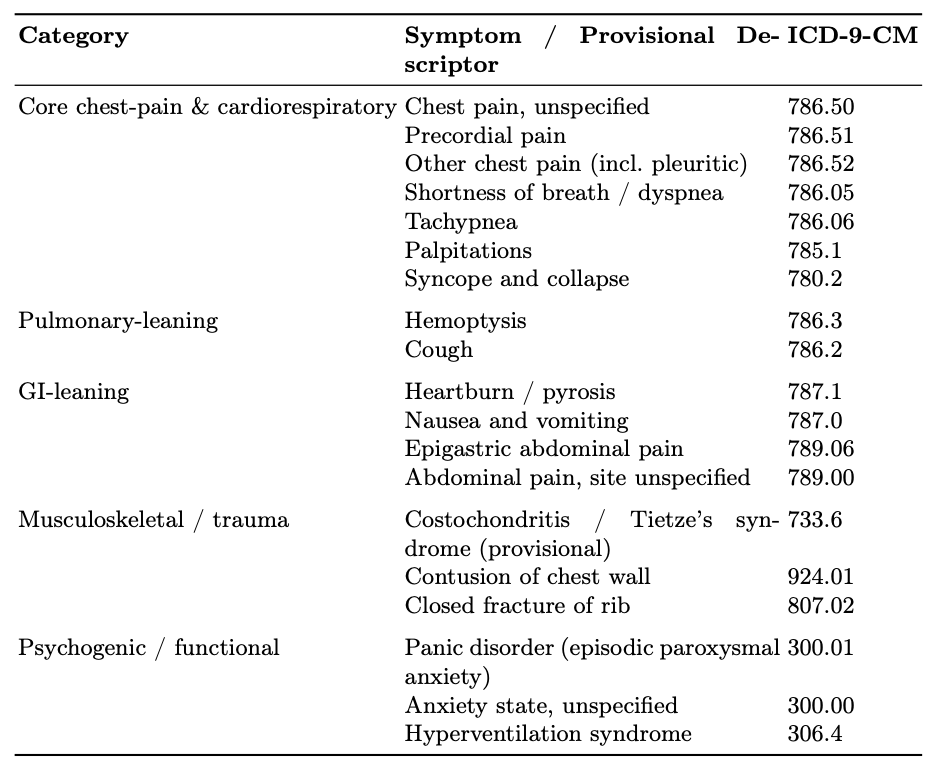}
    \caption{Initial set of diagnoses, notionally tied to associated diagnostic grouping}
    \label{fig:placeholder}
\end{figure}

\subsubsection{Final (Conclusive) ICD-9-CM Code Set — Specialist Gold Labels}
Figure 2 details the set of final, or target, Gold-Label diagnostic codes representing an ultimate diagnostic determination by a clinical expert. Patients must ultimately be diagnosed with one of these conditions for inclusion in the study cohort.

\small
\begin{figure}
    \centering
    \includegraphics[width=1\linewidth]{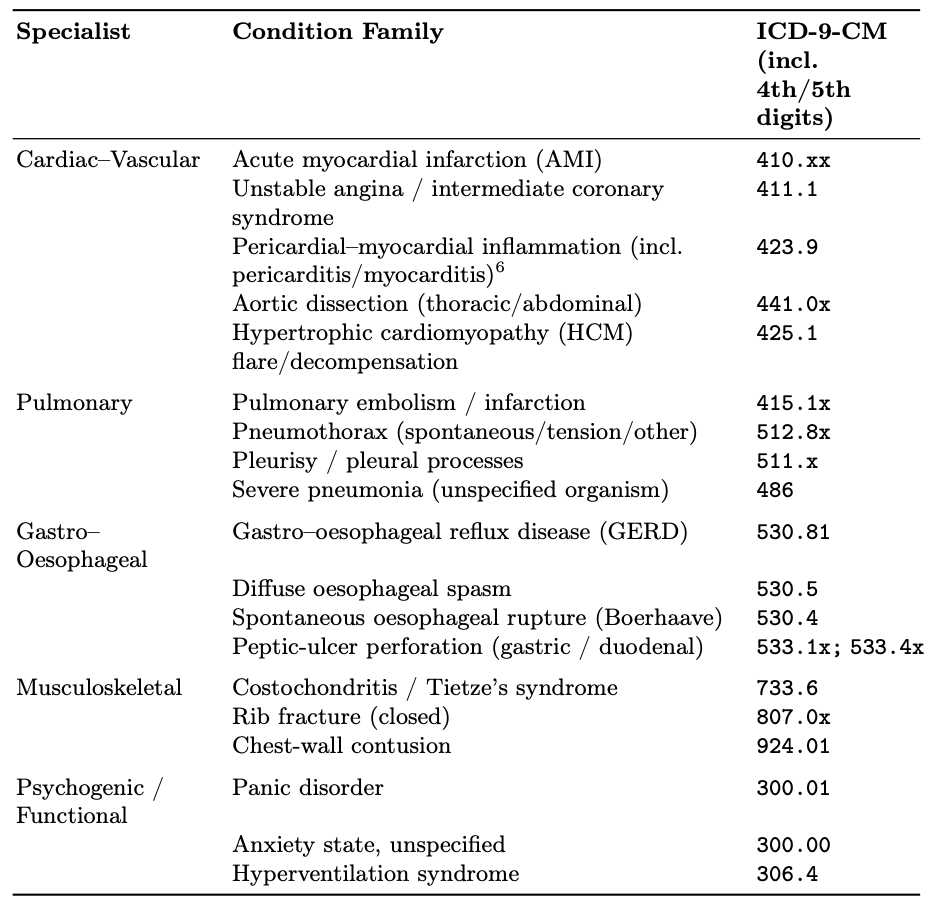}
    \caption{Set of final 'Gold Label' diagnostic codes, representing an ultimate diagnostic determination}
    \label{fig:placeholder}
\end{figure}
\normalsize

\noindent\textbf{Cohort composition note:}
The cohort is defined by the study's previously delineated  gold-label diagnostic families rather than the unconditional ED case mix; as a result, domain prevalence reflects selection criteria and may over-represent certain high-acuity presentations (e.g., cardiopulmonary). To mitigate dominance by any single domain, we report per-domain results in addition to macro-averaged metrics, and we treat evaluation on broader, less-selected cohorts as an external validation objective.

\subsubsection{Sequence construction}
Once eligible patients were determined, their medical histories were extracted and converted to token sequences using a fixed event-precedence policy, based foremost on chronological timing and then a predetermined hierarchy of token types: \texttt{DIAG}~$>$~\texttt{LAB}~$>$~\texttt{ORDER}, with explicit time-gap markers inserted between events when inter-event intervals exceed pre-specified thresholds (e.g., 1~hour, 6~hours) to preserve chronology. A final alphabetical ordering within each token-type is then done to ensure that batched sequences are ordered consistently across episodes (this ensures consistency in prediction accuracy given the singular 'next token' generation schema to the PRISM model). Each sequence is truncated or windowed to a maximum length of 512 tokens for an entire patient chronology, potentially spanning multiple admission episodes. The final (gold-label) diagnosis token is \emph{not} included on the input side for training tasks that predict it, preventing label leakage.

\subsubsection{Tokenization template and vocabulary}
We use a compact, hand-designed token schema with type prefixes to preserve semantics:\footnote{These templates are inherited from PRISM and reused verbatim. New domains add tokens but do not alter templates.}
\begin{itemize}
  \item \verb|[DIAG]|$\_$\verb|ICD9_|$\langle$code$\rangle$ for diagnoses (e.g., \verb|[DIAG]_ICD9_786.50|).
  \item \verb|[OBS]|$\_$\verb|LAB_|$\langle$test$\rangle$:$\langle$bin/value$\rangle$ for labs (e.g., \verb|[OBS]_LAB_TROP:HIGH|).\footnote{Continuous labs are discretized into clinically meaningful bins (e.g., LOW/NORMAL/HIGH/CRITICAL) defined a priori clinically-validated thresholds; raw numeric values are not emitted as free-text.}
  \item \verb|[ACTION]|$\_$\verb|ORD_|$\langle$order$\rangle$ for diagnostic orders (e.g., \verb|[ACTION]_ORD_ECG|).
  \item \verb|[GAP]|$\_$\verb|H|$\langle$k$\rangle$ for time-gap markers (e.g., \verb|[GAP]_H1| for $>$1~hour).
  \item Special sentinels: \verb|[BOS]|, \verb|[EOS]|, \verb|[PAD]|, \verb|[UNK]|.
\end{itemize}
Vocabulary growth is controlled by (i) whitelisting code families relevant to target presentations and (ii) merging infrequent variants into \verb|[UNK]| or higher-level bins (e.g., rare lab subtypes). This yields a compact vocabulary that supports low-latency inference while maintaining clinical interpretability.

\subsection{Model backbone (PRISM) Configuration and Training}
The baseline PRISM model is a decoder-only transformer with:
\begin{itemize}
  \item $L{=}6$ transformer blocks; model dimension $d_{model}{=}256$; MLP expansion $4d$; $n_{heads}{=}4$.
  \item Learned absolute positional embeddings; tied input/output token embeddings for parameter efficiency.
  \item Dropout $0.1$ in attention and MLP layers; layer normalization pre-attention.
\end{itemize}
The objective is next-token prediction over the event sequence (autoregressive cross-entropy). For temporal awareness, an auxiliary time-to-next-event head (Huber loss) can be added during pre-training; this head is discarded at inference.

\subsubsection{Optimization}
We train with AdamW (weight decay $0.01$), linear warmup over the first 5\% of steps followed by cosine decay to 10\% of the peak learning rate. Typical settings: batch size 64--128 sequences, peak LR in $1\text{e-}3$ to $2\text{e-}4$ depending on corpus size, gradient clipping at 1.0, early stopping on dev NDCG@3. Mixed-precision training is used when supported.

\subsubsection{Evaluation (PRISM baseline)}
Primary metrics for model performance were Top-1/Top-3 next-event recall, NDCG@k, and calibration (Brier score, reliability curves). For diagnosis-prediction ablations, AUROC/PR for specific gold-label families were measured. All metrics are computed per domain and macro-averaged to avoid dominance by common events.

\subsection{PRISM-Consult (panel of experts) Configuration and Training}

\begin{figure}[t]
\centering
\begin{tikzpicture}[
  box/.style={draw, rounded corners, align=center, inner sep=4pt, text width=0.90\columnwidth},
  arrow/.style={->, thick},
  note/.style={align=center, text width=0.90\columnwidth}
]

\node[box] (in) {Early ED prefix\\(first $K$ events)\\Tokenization + optional time features};

\node[box, below=4mm of in] (router) {Lightweight calibrated router\\$p(d \mid \text{prefix})$ over domains};

\node[box, below=4mm of router] (dispatch) {Dispatch policy\\Top-1 / Top-2 specialists\\Fail-open under low confidence};

\node[note, below=2mm of dispatch] (guard) {\footnotesize Safety guardrails: low confidence or danger thresholds $\Rightarrow$ consult-all};

\node[box, below=4mm of guard] (experts) {Domain specialists\\PRISM backbone + LoRA adapters\\Domain-aligned next-event / diagnosis-family scores};

\node[box, below=4mm of experts] (arb) {Deterministic arbitration / merge\\(e.g., Cardiac $>$ Pulmonary $>$ \dots)};

\node[box, below=4mm of arb] (out) {Outputs\\Predictions + uncertainty\\Audit log: routed domains, scores, thresholds};

\draw[arrow] (in) -- (router);
\draw[arrow] (router) -- (dispatch);
\draw[arrow] (dispatch) -- (experts);
\draw[arrow] (experts) -- (arb);
\draw[arrow] (arb) -- (out);

\end{tikzpicture}
\caption{PRISM-Consult end-to-end routing pipeline.}
\label{fig:prism-consult-pipeline}
\end{figure}

PRISM-Consult's framework extends PRISM by the expansion of the following elements:
\begin{enumerate}
  \item A \textbf{router} trained on the earliest events of each episode to emit one or more domain flags (Cardiac--Vascular, Pulmonary, Gastro--Oesophageal, Musculoskeletal, Psychogenic). Inputs are the first 2--5 coded events represented as bag-of-concepts TF--IDF features projected to a 256-dim space, optionally concatenated with simple temporal features (e.g., time since triage). The router is a logistic-regression classifier and a Platt scaling router with sigmoid outputs and class weighting to up-weight life-threatening domains.
  \item A set of \textbf{specialist models}---one per domain---initialized from the PRISM backbone and fine-tuned on domain-filtered corpora defined by the final gold-label codes specified above. To preserve parameter sharing and inference speed, each specialist uses low-rank adaptation (LoRA) on attention and feed-forward layers while keeping shared embeddings frozen; adapter ranks and learning rates are selected on a held-out dev set per domain.
  \item A \textbf{dispatch policy} at inference: if any life-threatening domain (e.g., AMI, PE, dissection) exceeds a calibrated threshold, a single high-priority specialist is invoked; otherwise the top-2 domains are consulted in parallel and their suggestions are merged by a deterministic arbitration layer (Cardiac $>$ Pulmonary $>$ Gastro $>$ Musculoskeletal $>$ Psychogenic).
\end{enumerate}

\subsubsection{Labeling surface (initial \& final codes)}
For each episode, only codes occurring \emph{before} the first definitive diagnosis are considered ``initial'' to prevent leakage; when multiple definitive diagnoses are present (e.g., AMI~+~PE), multi-label targets are assigned.

\subsubsection{Calibration, safety, and auditability}
All classifier outputs (router and specialists) were temperature-calibrated on a development fold using cross-entropy minimization. We logged router logits, specialist logits, and the final arbitration decision per episode to enable post hoc review. If the router emits uniformly low confidence (all domain probabilities below a safety threshold) or vitals cross pre-defined danger thresholds, the system fails open to parallel consultation of all specialists.

\subsubsection{Compute and reproducibility}
Experiments for the router were run on modern GPUs using deterministic seeds. Data preprocessing and training pipelines were version-controlled; model checkpoints, tokenizer vocabularies, and ICD-9 codebooks (both initial and final sets) are archived with SHA checksums.

\noindent\textbf{Reproducibility and artifacts.}
All experiments were run on \emph{NVIDIA A100 (Ampere) GPU with 40GB VRAM,} with 
with an \emph{8-core CPU} and \emph{83.5 GB of system RAM)} and software stack \emph{Linux Ubuntu, CUDA 12.x, NVIDIA proprietary driver}.

We provide the full training and evaluation pipeline, including preprocessing scripts, tokenizer vocabulary construction, fixed random seeds, and configuration files.
An accompanying public repository includes a minimal runnable notebook that reproduces key preprocessing and metric computations on a small illustrative example, enabling validation of the reported methodology.

\noindent\textbf{Code availability.}
Code for PRISM-Consult, including preprocessing, training, and evaluation scripts, along with a minimal runnable example notebook, is available in the PRISM repository.\footnote{\url{https://github.com/lmlevine/prism}}


\subsubsection{Overview of Light-Weight Router Design}
We model early Emergency Department (ED) episodes as a prefix time-series classification task. After the first $K$ coded events (default $K{=}5$), the router produces calibrated probabilities over five specialist domains: \emph{Cardiac--Vascular}, \emph{Pulmonary}, \emph{Gastro--Oesophageal}, \emph{Musculoskeletal}, and \emph{Psychogenic}. Episodes may be multi-label in principle, though the present corpus yielded disjoint (single-label) episodes. The router dispatches to one or more PRISM-Consult specialists under a safety-first policy.

\subsubsection{Data ingestion and harmonization}
Tokenized episodes reside in five domain directories (Cardiac, Pulmonary, Gastro--Oesophageal, Musculoskeletal, Psychogenic).

Each record is mapped to a canonical schema:
{\small
\[
\text{episode} \;=\; \bigl(\texttt{episode\_id},\; \mathbf{e}=[e_1,\ldots,e_{L}],\; \texttt{time\_feats}\in\mathbb{R}^{\le 2}\bigr),
\]
}
\normalsize
where optional \texttt{time\_feats} contain simple scalars (minutes-to-first-order; max inter-event gap). Episodes with the same \texttt{episode\_id} across directories are merged (longest token list retained; non-empty \texttt{time\_feats} preferred), and a 5-dim multi-hot label $\mathbf{y}\in\{0,1\}^5$ indicates domain membership. In this study, directories were disjoint by construction, so $\|\mathbf{y}\|_0{=}1$.

\subsubsection{Proportional sampling}
Let $\mathcal{E}=\{(\mathbf{e}_i,\mathbf{y}_i)\}_{i=1}^{N}$ denote the merged episode table. To obtain a target size $T$ while preserving domain mixture, we compute per-domain prevalence $p_d$ on $\mathcal{E}$ and allocate quotas $q_d\approx p_d T$. We first fill quotas with domain-exclusive episodes, then top up from remaining sets (including potential multi-label cases), avoiding duplicates; if fewer than $T$ remain after deduplication, we top up uniformly at random. Setting $T{=}0$ disables sampling.

\subsubsection{Prefix expansion (anytime supervision)}
To enable predictions after each early event, we expand episodes into prefixes of lengths $\ell \in \{1,\dots,\min(K,L_i)\}$:
\[
\mathcal{P} \;=\; \bigl\{(i,\ell,\;\mathbf{e}_{i,1:\ell},\;\mathbf{y}_i,\;w_\ell=\ell/K)\bigr\}.
\]
Each prefix inherits the final label $\mathbf{y}_i$ but only uses the first $\ell$ tokens as input, preventing label leakage. We optionally use $w_\ell$ as a sample weight to mildly emphasize later (more informative) prefixes.

\subsubsection{Featurization: TF--IDF $\rightarrow$ SVD}
We join prefix tokens with spaces to form a short ``document'' and compute 1--2-gram TF--IDF with \texttt{min\_df=2}. A 256-dim truncated SVD yields a compact vector $\mathbf{z}_{i,\ell}\in\mathbb{R}^{256}$; optional \texttt{time\_feats} ($\le$2 scalars) are concatenated to form $\tilde{\mathbf{z}}_{i,\ell}$.

\subsubsection{Router model and calibration}
We train one-vs-rest logistic regression heads (saga, L2, $C{=}2.0$, \texttt{max\_iter}=3000) for the five domains on the prefix table $\mathcal{P}$, using stratified patient-level splits (70/10/20 train/dev/test on the indicator of any positive label). Probabilities are calibrated per head on the development split with Platt scaling (3-fold). Let $\hat{p}_d(\tilde{\mathbf{z}})$ be the calibrated probability for domain $d$.

\subsubsection{Routing policy and threshold tuning}
At inference, with $\hat{\mathbf{p}}=\{\hat{p}_d\}_{d=1}^{5}$:
\begin{enumerate}
  \item If a life-threatening domain (Cardiac, Pulmonary) satisfies $\hat{p}_d \ge \tau_{\text{hi}}$, route \emph{top-1} (the argmax).
  \item Else if $\max_d \hat{p}_d \ge \tau_{\text{lo}}$, route \emph{top-2}.
  \item Else, \emph{fail-open} to all five experts.
\end{enumerate}

We grid-search $(\tau_{\text{hi}},\tau_{\text{lo}})$ on the development split to minimize expected experts per episode subject to a safety constraint on life-threatening recall (Cardiac$\lor$Pulmonary). Unless otherwise noted, we target $\ge 0.98$ on development; if no grid point satisfies the constraint, we select the point with maximal life-threat recall (tie-break: lower compute).

\subsubsection{Safety behavior}
If calibrated probabilities are uniformly low (i.e., $\max_d \hat{p}_d {<} 0.25$) or if triage vitals cross pre-defined danger thresholds, the router fails open to all experts and emits an audit record (raw logits, thresholds, selected route). Temperature scaling is re-checked on each model refresh to maintain probability fidelity.

\subsubsection{Evaluation}
\emph{Discrimination.} ROC-AUC and PR-AUC are reported per domain on the held-out test set using calibrated probabilities.
\paragraph{Routing quality} With routed set $R_i$ and truth $Y_i$ for episode $i$:
\[
\text{Recall}_{\text{any}}=\tfrac{1}{|\mathcal{I}|}\sum_{i\in\mathcal{I}}\mathbf{1}[R_i\cap Y_i\neq\varnothing],\quad
\]
\[
\text{Recall}_{\text{all}}=\tfrac{1}{|\mathcal{I}|}\sum_{i\in\mathcal{I}}\mathbf{1}[Y_i\subseteq R_i].
\]
Life-threat recall is computed on:

$\{i: Y_i\cap\{\text{Cardiac},\text{Pulmonary}\}\neq\varnothing\}$

and requires $R_i$ to include at least one of these domains.
\paragraph{Compute proxy and latency} We report $\mathbb{E}[|R|]=\tfrac{1}{N}\sum_i |R_i|$ and an estimated latency $L_i=L_{\text{router}}+\sum_{d\in R_i}L_d$ using fixed per-expert times.
\paragraph{Anytime curves} Metrics are stratified by prefix length $\ell=1,\dots,K$ to quantify earliness.

\paragraph{Anytime performance} To quantify earliness, we repeat the above at each prefix length $\ell \in \{1,\dots,K\}$, plotting discrimination and routing recalls as functions of $\ell$.

\subsubsection{Ablations and robustness checks}
We assess: (i) $K \in \{2,3,5\}$; (ii) text-only vs.\ text+time features; (iii) uncalibrated vs.\ calibrated probabilities; and (iv) alternative linear heads (linear SVM with Platt). As sanity baselines we compare against \emph{consult-all}, \emph{fixed Cardiac+Pulmonary}, and a single generalist PRISM variant (no routing).

\section{Results}

\subsection{Specialist Model Training \& Performance}

\subsubsection{Cohorts and training setup}
A total of $N{=}20{,}436$ Emergency Department (ED) episodes met all inclusion criteria for PRISM\mbox{-}Consult. Domain-eligible cohorts are not disjoint (multi-label episodes may appear in multiple domain pools). Where noted, some domains were \emph{scoped} to a capped training subset to ensure compute parity and class balance. All specialists inherit the PRISM backbone and tokenization; optimization and early-stopping follow the protocol in the Methods section. Unless otherwise stated, training used the default causal language modeling loss of autoregressive cross-entropy) with mixed precision.

\subsubsection{Domain-specific outcomes (Specialist Model Set Performance)}
Each specialist exhibited stable convergence with steadily improving validation loss across epochs, consistent with the baseline PRISM behavior on cardiac timelines. Table~\ref{tab:spec-results} reports the best development loss, derived perplexity ($\mathrm{PPL}=\exp(\mathrm{loss})$), and percentage reduction from epoch~1 to the best epoch.

\begin{figure}

    \includegraphics[width=1\linewidth]{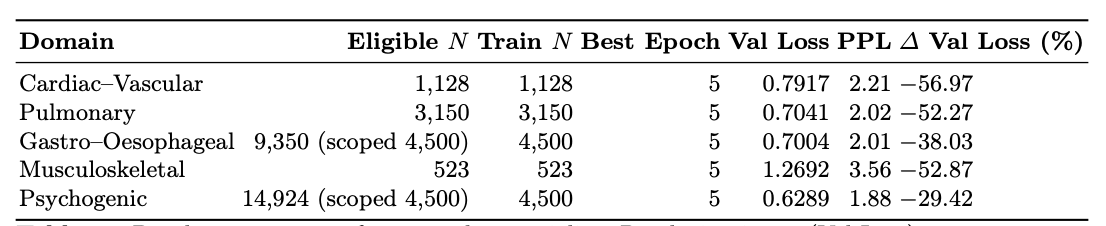}
    \caption{Development-set performance by specialist model. Perplexity is $\exp(\text{Val Loss})$. $\Delta$~Val Loss is the percentage reduction from epoch~1 to the best epoch. Scoped domains used capped training sets for balance.}
    \label{fig:placeholder}
\end{figure}
\label{tab:spec-results}

\paragraph{Cardiac--Vascular (AMI/UA/pericardial/aortic/HCM)}
Validation loss decreased from $1.8397$ at epoch~1 to $0.7917$ at epoch~5 ($-56.97\%$; $\mathrm{PPL}{=}2.21$), closely mirroring the original PRISM cardiac behavior and supporting the transferability of the backbone.

\paragraph{Pulmonary (PE/pneumothorax/pleura/pneumonia)}
Validation loss improved from $1.4753$ to $0.7041$ by epoch~5 ($-52.27\%$; $\mathrm{PPL}{=}2.02$), indicating a well-captured structure in early pulmonary presentations.

\paragraph{Gastro--Oesophageal (GERD/spasm/Boerhaave/ulcer perforation)}
On a scoped training pool of 4{,}500 episodes, validation loss fell from $1.1303$ to $0.7004$ ($-38.03\%$; $\mathrm{PPL}{=}2.01$) by epoch~5, with steady epoch-on-epoch gains and no signs of overfitting.

\paragraph{Musculoskeletal (costochondritis/rib injury/chest-wall contusion)}
Despite the smallest cohort ($N{=}523$), validation loss dropped from $2.6927$ to $1.2692$ ($-52.87\%$; $\mathrm{PPL}{=}3.56$). The higher perplexity is consistent with data scarcity and etiologic heterogeneity; nevertheless, the relative improvement and stable trajectory indicate learnability under limited data.

\paragraph{Psychogenic (panic/anxiety/hyperventilation)}
Training on a scoped subset of 4{,}500 episodes converged from $0.8910$ to $0.6289$ ($-29.42\%$; $\mathrm{PPL}{=}1.88$), the lowest perplexity among the specialists, reflecting a relatively compact symptom-to-diagnosis mapping for this domain.

\begin{figure}
    \centering
    \includegraphics[width=1\linewidth]{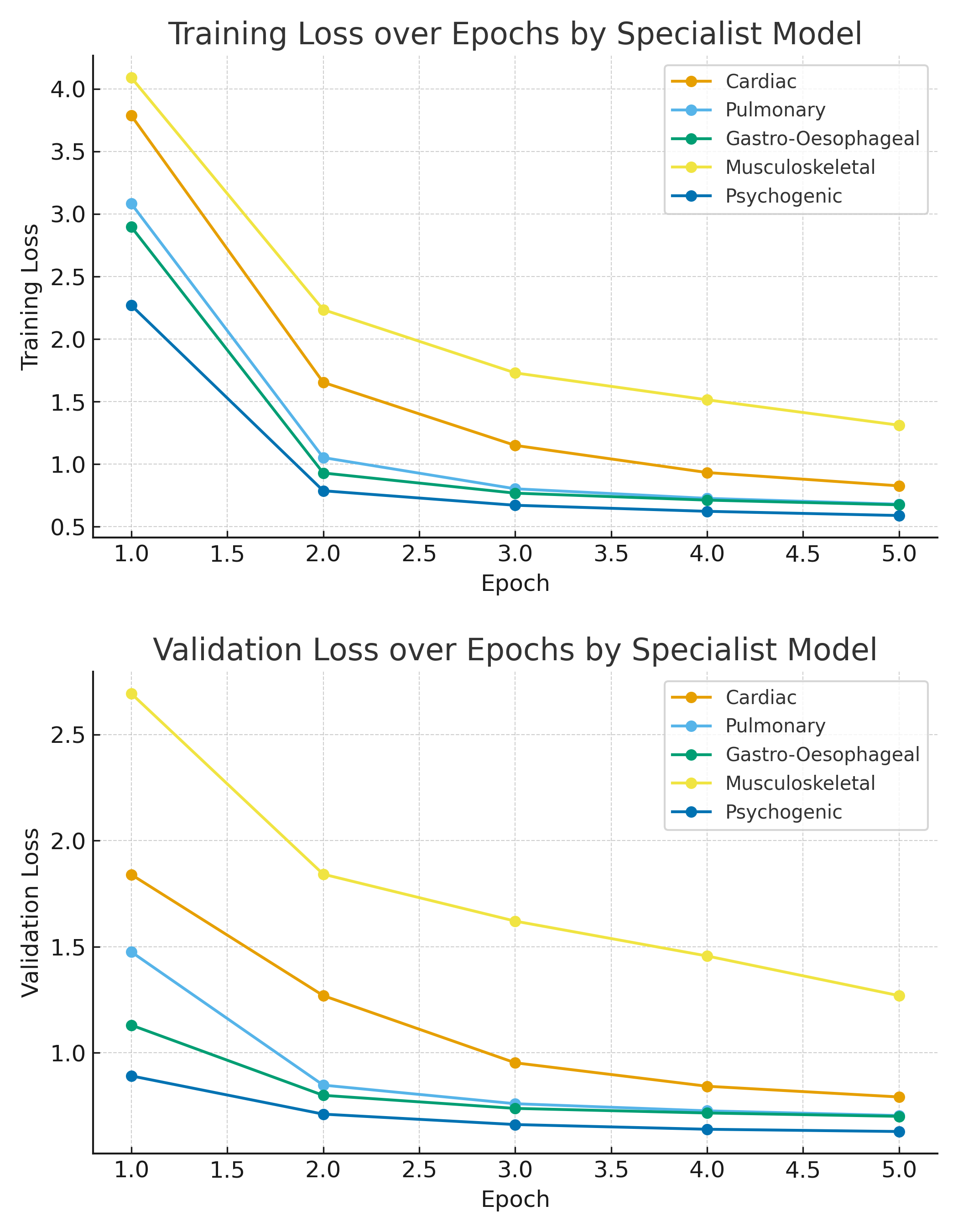}
    \caption{Cross-specialist Training Loss over epochs for both training and validation cohorts}
    \label{fig:placeholder}
\end{figure}

\subsubsection{Cross-domain interpretation}
Figure 4 visualizes the training of all specialist models across epochs. All five specialists converge to low development perplexities (three near ${\sim}2.0$ and one below $2.0$), with smooth validation curves and substantial loss reductions from epoch~1 (Table~\ref{tab:spec-results}). This \emph{cross-disciplinary consistency}---achieved without altering tokenization, model size, or optimization hyperparameters---validates our central design choice: the compact PRISM backbone, when routed to domain-consistent corpora, yields strong and stable learning dynamics beyond its original cardiac scope. The musculoskeletal model’s higher perplexity likely reflects limited sample size and broader label variance; we anticipate improvements with targeted data augmentation and modest adapter-rank increases.

\subsubsection{Training curves (dev) and calibration}
For each specialist, validation loss decreased monotonically through epoch~5 with no divergence from training loss, suggesting adequate regularization (dropout 0.1; early stopping at epoch~5). Final checkpoints correspond to the minima reported above. Temperature scaling for probabilistic outputs is fit on the development fold, with reliability plots to be included in the appendix.

\subsection{Router Training Results}
\label{sec:results}

\subsubsection{Cohort and training rows}
Across domain directories we ingested $N{=}13{,}801$ unique episodes: Cardiac $1{,}128$ (8.2\%), Pulmonary $3{,}150$ (22.8\%), Gastro--Oesophageal $4{,}500$ (32.6\%), Musculoskeletal $523$ (3.8\%), Psychogenic $4{,}500$ (32.6\%). Episodes were single-label by design (no cross-directory duplicates). Prefix expansion with $K{=}5$ yielded $69{,}005$ training/evaluation rows.

\subsection{Cohort and Routing Policy Summary}
\begin{figure}
    \centering
    \includegraphics[width=1\linewidth]{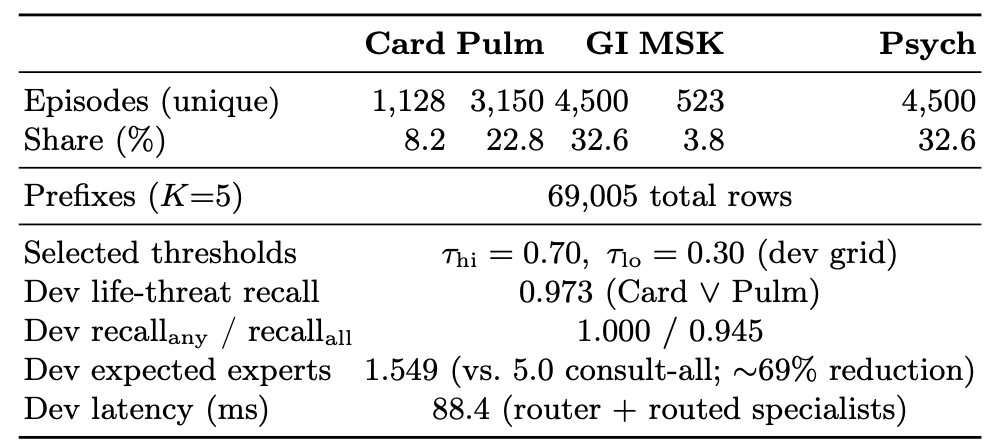}
    \caption{Light-Weight Router Training Cohort and Overall Results Summary}
    \label{fig:placeholder}
\end{figure}

\subsubsection{Selected thresholds and routing mix}
Fig 5. details specific thresholds and overall results mix. The development grid selected $(\tau_{\text{hi}},\tau_{\text{lo}})=(0.70,\,0.30)$ under our objective. With these thresholds, the router predominantly returned \emph{top-1} or \emph{top-2} routes; fail-open was rare. The expected consulted experts per episode on test was $1.565$, a $\sim\!69\%$ reduction versus consult-all (5 experts).

\subsection{Discrimination (per-domain; macro-averaged)}
Fig. 6 details domain specific descriminitive results, along with macro-averaged results across domains for both the development and testing cohorts.
\begin{figure}
    \centering
    \includegraphics[width=1\linewidth]{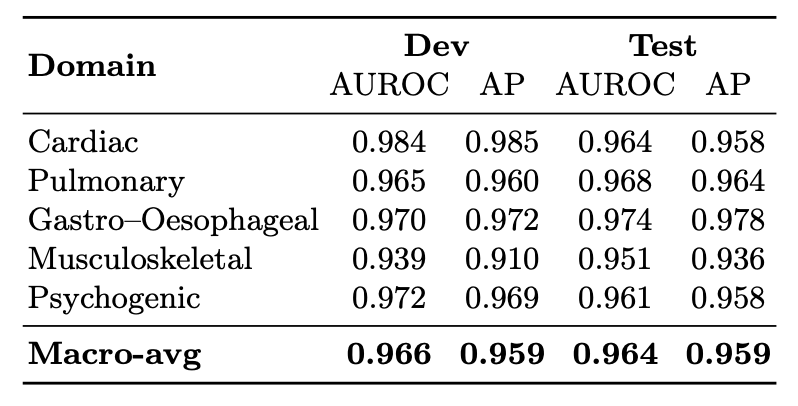}
    \caption{Domain-specific discriminitive results}
    \label{fig:placeholder}
\end{figure}

\subsubsection{Safety and routing quality}
On the development split, life-threatening recall (Cardiac$\lor$Pulmonary) was $0.973$; on test it was $0.965$. Both \emph{Recall\textsubscript{any}} metrics were $1.000$ (the routed set always included a correct domain), and \emph{Recall\textsubscript{all}} was $0.945$ (dev) and $0.942$ (test). Estimated latencies, using the fixed per-expert constants, were $88.4$\,ms (dev) and $89.5$\,ms (test). Fig 8 details the overall end-to-end PRISM-Consult system summary performance.

\begin{figure}
    \centering
    \includegraphics[width=1\linewidth]{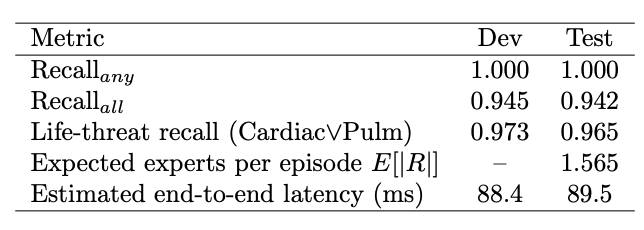}
    \caption{End-to-end PRISM-Consult system summary (routing safety and efficiency)}
    \label{fig:placeholder}
\end{figure}


\section{Discussion}

\subsection{Cross-disciplinary performance of PRISM}
PRISM-Consult extends the original PRISM design from a cardiology-focused study to a routed, multi-specialist system while preserving the compact backbone and token template. Empirically, all five specialists trained with the same optimization recipe and achieved smooth, monotonic declines in development loss with low final perplexities (near $\sim$2.0 for Cardiac, Pulmonary, Gastro–Oesophageal, and Psychogenic; higher but improving for Musculoskeletal). This cross-disciplinary consistency, obtained without increasing model width/depth or altering tokenization, indicates that the PRISM backbone transfers effectively when exposed to domain-consistent corpora and calibrated with light routing. That is, a single representational space and schema can support heterogeneous organ-system reasoning as long as data partitioning and supervision align with clinical provenance. 

Additionally the router’s near-perfect \textit{Recall\textsubscript{any}} and high \textit{Recall\textsubscript{all}} further suggest that early tokens carry enough discriminative signal to select the appropriate specialist(s) reliably, even under strict latency constraints.

\subsection{Cost-sensitive clinical utility (future work)}
While we report standard discrimination and ranking metrics, real clinical deployment is cost-sensitive: the consequence of a miss depends on severity (e.g., failing to identify a time-critical cardiopulmonary presentation) and the consequence of ``over-consulting'' is often primarily resource or latency cost. A principled extension is to report an expected-utility (or expected-cost) metric defined over the routed specialist set. For episode $i$ with true domain set $Y_i$ and routed set $R_i$, one can define:
\begin{equation}
C_i = \sum_{d \in Y_i \setminus R_i}\lambda_d \;+\; \gamma\sum_{d \in R_i \setminus Y_i} 1,
\end{equation}
where $\lambda_d$ is higher for life-threatening domains and $\gamma$ captures the comparatively lower cost of over-routing. This yields an explicit compute--safety Pareto curve as router thresholds vary. In the current study, we did not retain per-episode routed-set outputs for all runs, which prevents retrospective computation of such utility-based metrics from aggregate results alone. We therefore include this evaluation as a priority direction for future work and as a recommended standard for high-stakes clinical sequence prediction.

\subsection{Next steps: validation and extension}
We outline three directions to harden and extend this framework:
\begin{enumerate}
  \item \textbf{External and temporal validation.} Replicate training/evaluation on (i) a second site with distinct coding habits and lab panels; (ii) a temporally held-out slice to assess drift. 
  \item \textbf{Safety-first routing refinements.} Enforce a hard development constraint of life-threatening recall $\geq 0.98$ via (a) asymmetric thresholds (lower $\tau_{\mathrm{hi}}$ for Cardiac/Pulmonary than for other domains), (b) a guardrail that includes both Cardiac and Pulmonary whenever $\max(\hat p_{\mathrm{card}},\hat p_{\mathrm{pulm}})\geq \tau_{\mathrm{lo}}^{\mathrm{life}}$, and (c) per-head isotonic calibration when under-confidence is detected near decision boundaries. Re-report the compute–safety Pareto frontier (expected experts vs.\ life-threat recall).
  \item \textbf{Broader coverage and richer inputs.} Add specialists (e.g., aortic catastrophes vs.\ general vascular, neurologic mimics) as new gold-label families mature; incorporate small, domain-agnostic time features (minutes-to-first-order; max inter-event gap) that are already supported by the router. Where multi-label episodes exist (e.g., AMI+PE), explicitly train/evaluate multi-label routing quality (\textit{Recall\textsubscript{all}} at top-$k$) and measure arbitration behavior.
\end{enumerate}
Operationally, we recommend prefix-stratified audits (metrics by $\ell=1..K$), route-mix telemetry (top-1/top-2/fail-open), and episode-level audit logs (router/specialist logits, thresholds, decision) to support prospective QA and IRB-facing safety reviews.

\subsection{Limits}
Three significant limitations temper interpretation: 
\paragraph{Multi-label handling (limitation and evaluation plan)}
First, the present corpus is effectively single-label across directories, so the multi-label routing regime was not strongly exercised. We therefore treat multi-label arbitration as a priority validation objective. In future work we will evaluate confirmed multi-diagnosis cases (e.g., AMI+PE) and report multi-label routing quality (e.g., Recallall@k / top-$k$ coverage, calibration by prefix length), along with structured audits of arbitration outcomes and fail-open behavior in safety-critical presentations.

\paragraph{Limited Clinical Domains and Skewed Patient-Mix}
Second, this study evaluates only five specialist families selected to support clear gold-labeling and clinically meaningful ED coverage; this does not represent the full breadth of ED consult services. In addition, cohort composition reflects the selected diagnostic families and may be skewed toward higher-acuity presentations resulting in class imbalance (e.g., the relatively small Musculoskeletal pool) likely contributes to higher perplexity and calibration variance for that head; targeted sampling, modest adapter rank increases, and data augmentation (token normalization for near-synonymous events) are straightforward mitigations for class imbalance, and future work should also aim to expand the specialist panel (e.g., neurologic, infectious/sepsis, renal/metabolic, toxicologic/trauma) and evaluate scalability of routing and arbitration as the panel grows, and (ii) validate generalization on broader, less-selected cohorts and across sites/time periods to better reflect real-world ED case mix. 
\paragraph{Fixed Positional Embeddings may Prove too Rigid for Model Adaptability}
Third, learned absolute positional embeddings fix the maximum context length and may not extrapolate; if longer horizons are needed, rotary or ALiBi encodings can be substituted with minimal disruption to the rest of the stack. 

Additionally, we note that threshold selection in the current run favored a near-constraint point (life-threat recall $<0.98$ on test); future reports should hard-enforce prespecified safety constraints and present confidence intervals via bootstrap over episodes.

Finally, to better contextualize model accuracy and efficiency, PRISM-Consult should be benchmarked against (i) classical classifiers (e.g., XGBoost) trained on the same early-prefix feature representation, and (ii) compact encoder models (e.g., ClinicalBERT/BioBERT-style) trained to predict domain/diagnosis from prefix tokens.

\section{Conclusion}
PRISM-Consult operationalizes a clinician-aligned \emph{panel-of-experts} by pairing a calibrated, light-weight router with parameter-efficient PRISM specialists trained on domain-consistent corpora. Using only the earliest tokens, the router attains high coverage and substantial compute savings relative to consult-all, while specialists exhibit smooth, low-perplexity convergence across disparate organ systems—evidence that the PRISM backbone generalizes beyond its original cardiac scope. With minor policy and calibration refinements to meet strict life-threat recall targets, the framework provides an auditable, low-latency pathway to deployable clinical decision support that routes each episode to the right expert at the right time.


\vspace{12pt}


\begin{thebibliography}{99}

\bibitem{levine2025prism}
L.~Levine, J.~Santerre, \emph{et al.}, 
''PRISM: A Transformer-based Language Model of Structured Clinical Event Data,'' \emph{arxiv.org/abs/2506.11082}, 2025.

\bibitem{lee2019biobert}
J.~Lee, W.~Yoon, S.~Kim, \emph{et al.}, ``BioBERT: a pre-trained biomedical language representation model for biomedical text mining,'' \emph{arXiv:1901.08746}, 2019.

\bibitem{huang2019clinicalbert}
K.~Huang, A.~Altosaar, D.~R.~R. Ranganath, ``ClinicalBERT: Modeling clinical notes and predicting hospital readmission,'' \emph{arXiv:1904.05342}, 2019.

\bibitem{gu2021pubmedbert}
Y.~Gu, R.~Tinn, H.~Cheng, \emph{et al.}, ``Domain-Specific Language Model Pretraining for Biomedical NLP,'' in \emph{ACL}, 2021.

\bibitem{luo2022biogpt}
R.~Luo, L.~Sun, Y.~Xia, \emph{et al.}, ``BioGPT: Generative Pre-trained Transformer for Biomedical Text Generation and Mining,'' \emph{arXiv:2210.10341}, 2022.

\bibitem{yang2022gatortron}
X.~Yang, A.~W.~S. Zhang, Y.~Bian, \emph{et al.}, ``A large language model for electronic health records,'' \emph{npj Digital Medicine} 5, 194 (2022).

\bibitem{peng2023gatortrongpt}
C.~Peng, X.~Yang, Y.~Bian, \emph{et al.}, ``A study of generative large language model for medical AI,'' \emph{npj Digital Medicine} 6, 205 (2023).

\bibitem{singhal2023largelanguage}
K.~Singhal, S.~Azizi, T.~Tu, \emph{et al.}, ``Large language models encode clinical knowledge,'' \emph{Nature} 620, 172–180 (2023).

\bibitem{medpalm2}
K.~Singhal, S.~Azizi, E.~K.~Lai, \emph{et al.}, ``Med-PaLM 2: Towards expert-level medical question answering with LLMs,'' \emph{arXiv:2305.09617}, 2023.

\bibitem{lepikhin2020gshard}
D.~Lepikhin, H.~Lee, Y.~Xu, \emph{et al.}, ``GShard: Scaling giant models with conditional computation and automatic sharding,'' \emph{arXiv:2006.16668}, 2020.

\bibitem{fedus2021switch}
W.~Fedus, B.~Zoph, N.~Shazeer, ``Switch Transformers: Scaling to trillion parameter models with simple and efficient sparsity,'' \emph{arXiv:2101.03961}, 2021.

\bibitem{expertchoice}
Z.~Yanqi, L.~Tao, \emph{et al.}, ``Mixture-of-Experts with Expert Choice Routing,'' Google Research Blog (Nov.~2022), NeurIPS 2022 spotlight.

\bibitem{routellm-openreview}
I.~Ong, A.~Huang, P.~Lu, \emph{et al.}, ``RouteLLM: Learning to Route LLMs with Preference Data,'' \emph{OpenReview}, 2024.

\bibitem{eagle-router}
Z.~Zhao, F.~Huang, Y.~Xu, \emph{et al.}, ``Eagle: Efficient Training-Free Router for Multi-LLM Inference,'' NeurIPS ML For Systems Workshop, 2024.

\bibitem{moa2024}
J.~Wang, J.~Wang, B.~Athiwaratkun, C.~Zhang, J.~Zou, ``Mixture-of-Agents Enhances Large Language Model Capabilities,'' \emph{arXiv:2406.04692}, 2024.

\bibitem{masanneck2024triage}
L.~Masanneck, M.~G{\"u}nther, S.~K{\"o}rber, \emph{et al.}, ``Triage Performance Across Large Language Models and Healthcare Professionals,'' \emph{J Med Internet Res} 26:e53297 (2024).

\bibitem{jama2024diagnostic}
S.~R.~Ranji, A.~F.~Hernandez-Boussard, ``Large Language Models—Misdiagnosing Diagnostic Reasoning?,'' \emph{JAMA Network Open} 7(10):e2433381 (2024).

\bibitem{wasylewicz2018cds}
A.~T.~M.~Wasylewicz, M.~W.~M.~Jaspers, ``Clinical Decision Support Systems,'' in \emph{Health Informatics: eHealth} (NIH/NCBI Bookshelf), 2018.


\bibitem{johnson2023mimiciv}
A.~E.~W.~Johnson, L.~Bulgarelli, L.~Shen, A.~Gayles, A.~Shammout, S.~Horng, T.~J.~Pollard, B.~Moody, B.~Gow, L.-w.~H.~Lehman, L.~A.~Celi, and R.~G.~Mark,
``MIMIC-IV, a freely accessible electronic health record dataset,''
\emph{Scientific Data} 10(1), 1--18 (2023).
doi:10.1038/s41597-022-01899-x.


\bibitem{sadi2024ddxplus}
F.~Sadi~et~al.,
``DDXPlus: A New Benchmark Dataset for Differential Diagnosis,''
\emph{Journal of Medical Internet Research} (2024).
\bibitem{yang2023ddxt}
J.~Yang~et~al.,
``DDxT: A Generative Transformer for Differential Diagnosis Generation,''
\emph{Proceedings of the Conference on Health AI (CHAI)} (2023).
\bibitem{yuan2021aarlc}
M.~Yuan and L.~Yu,
``AARLC: Adaptive Action Reinforcement Learning for Clinical Diagnosis,''
\emph{Artificial Intelligence in Medicine} 118 (2021).
\bibitem{tchango2022basd}
F.~Tchango~et~al.,
``BASD: Bayesian-Assisted Sequential Diagnosis,''
\emph{Medical Informatics Journal} (2022).
\bibitem{chen2022dxformer}
S.~Chen~et~al.,
``DxFormer: A Transformer-Based Model for Medical Dialogue and Diagnosis,''
\emph{Proceedings of the AAAI Conference on Artificial Intelligence} 36 (2022).
\bibitem{yang2024adaptivetopk}
B.~Yang~et~al.,
``Adaptive Top-K Reinforcement Learning for Efficient Clinical Diagnostics,''
\emph{Nature Biomedical Engineering} (2024).
\bibitem{guan2021bayesian}
X.~Guan and C.~Baral,
``Bayesian Experimental Design for Sequential Diagnosis,''
\emph{Artificial Intelligence Journal} 298, 103503 (2021).

\bibitem{9380633}
R.~F.~Mansour, A.~E.~Amraoui, I.~Nouaouri, V.~G.~Díaz, D.~Gupta, and S.~Kumar,
``Artificial Intelligence and Internet of Things Enabled Disease Diagnosis Model for Smart Healthcare Systems,''
\emph{IEEE Access} 9, 45137--45146 (2021).
doi:10.1109/ACCESS.2021.3066365.

\bibitem{autoddx_2024}
Abu Adnan Sadi and Mohammad Ashrafuzzaman Khan and Lubaba Binte Saber,
''Automatic Differential Diagnosis Using Transformer‐Based Multi-Label Sequence Classification,''
\emph{arXiv preprint} (2024).
arXiv:2408.15827.

\bibitem{ethos_2024}
P.~Renc, Y.~Jia, A.~E.~Samir, J.~Was, Q.~Li, D.~W.~Bates, and A.~Sitek,
``Zero shot health trajectory prediction using transformer,''
\emph{npj Digital Medicine} 7(1) (2024).
doi:10.1038/s41746-024-01235-0.

\bibitem{medalbert_2024}
Anonymous,
``Transformer-Based Deep Learning Model for the Diagnosis of Lung Cancer in Primary Care,''
\emph{npj Digital Medicine} (2024).
\bibitem{lotusai_2025}
Lotus~Health~AI,
``Transforming Disease Prediction with {LotusAI-Predict}: A Fine-Tuned {LLaMA} Model,''
\emph{Lotus Health AI Blog} (2025).
URL: \url{https://lotushealth.ai/blog/transforming-disease-prediction-with-lotusai-predict}.
Accessed: 11~Apr.~2025.
\bibitem{dpss_2020}
Y.~Zhang, L.~Chen, and D.~Ghosh,
``Diagnostic Prediction with Sequence-of-Sets Representation Learning for Clinical Events,''
\emph{Proceedings of the ACM Conference on Health, Inference, and Learning} (2020).
\bibitem{setor_2023}
Anonymous,
``Sequential Diagnosis Prediction with Transformer and Ontological Representation,''
\emph{IEEE Journal of Biomedical and Health Informatics} (2023).
\bibitem{jaakkola1999variational}
T.~S.~Jaakkola and M.~I.~Jordan,
``Variational probabilistic inference and the QMR-DT network,''
\emph{Journal of Artificial Intelligence Research} 10, 291--322 (1999).
\bibitem{young2022empirical}
Z.~Young and R.~Steele,
``Empirical evaluation of performance degradation of machine learning-based predictive models--A case study in healthcare information systems,''
\emph{International Journal of Information Management Data Insights} 2(1), 100070 (2022).

\bibitem{combi2009temporal}
C.~Combi, M.~Gozzi, B.~Oliboni, J.~M.~Juarez, and R.~Marin,
``Temporal similarity measures for querying clinical workflows,''
\emph{Artificial Intelligence in Medicine} 46(1), 37--54 (2009).

\bibitem{zhang2015cowpath}
Y.~Zhang, R.~Padman, and N.~Patel,
``Paving the COWpath: Learning and visualizing clinical pathways from electronic health record data,''
\emph{Journal of Biomedical Informatics} 58, 186--197 (2015).

\bibitem{aspland2021modified}
E.~Aspland, P.~R.~Harper, D.~Gartner, P.~Webb, and P.~Barrett-Lee,
``Modified Needleman--Wunsch algorithm for clinical pathway clustering,''
\emph{Journal of Biomedical Informatics} 115, 103668 (2021).

\bibitem{zhong2022clustering}
H.~Zhong, G.~Loukides, and S.~P.~Pissis,
``Clustering demographics and sequences of diagnosis codes,''
\emph{IEEE Journal of Biomedical and Health Informatics} 26(5), 2351--2359 (2022).

\bibitem{arnolds2018improving}
I.~V.~Arnolds and D.~Gartner,
``Improving hospital layout planning through clinical pathway mining,''
\emph{Annals of Operations Research} 263(1), 453--477 (2018).

\bibitem{budde2017road}
C.~E.~Budde, P.~R.~D'Argenio, R.~E.~Monti, M.~D.~Lee, L.~Rodr\'iguez, and N.~Wolovick,
``The Road from Stochastic Automata to the Simulation of Rare Events,''
in \emph{ModelEd, TestEd, TrustEd}, eds. R.~Langerak, J.-P.~Katoen, and A.~Rensink,
\emph{Lecture Notes in Computer Science}, Springer, pp.~276--294 (2017).
doi:10.1007/978-3-319-68270-9\_14.


\end{thebibliography}
\end{document}